%% file: main.tex
\documentclass[10pt,twocolumn,letterpaper]{article}

\usepackage[review,algorithms]{wacv}      

\usepackage{booktabs}
\usepackage{array}     
\usepackage{makecell}  
\newcolumntype{M}[1]{>{\centering\arraybackslash}p{#1}} 

\usepackage{graphicx}
\usepackage{subcaption} 


\definecolor{wacvblue}{rgb}{0.21,0.49,0.74}
\usepackage[pagebackref,breaklinks,colorlinks,allcolors=wacvblue]{hyperref}

\def\wacvPaperID{13} 
\def\confName{WACV}
\def\confYear{2026}

\title{Subimage Overlap Prediction: Task-Aligned Self-Supervised Pretraining For Semantic Segmentation In Remote Sensing Imagery}

\author{Lakshay Sharma\\
Instacart, New York University\\
New York, NY\\
{\tt\small lakshay.sharma@cims.nyu.edu}
\and
Alex Marin\\
Thomson Reuters, University of Washington\\
Seattle, WA\\
{\tt\small amarin@uw.edu}
}

\begin{document}
\maketitle
\input{sec/0_abstract}    
\input{sec/1_intro}
\input{sec/2_related_work}
\input{sec/3_subimage_overlap_pretraining}
\input{sec/4_downstream}
\input{sec/4_1_po-downstream-ablation}
\input{sec/4_2_varying-downstream-data}
\input{sec/4_3_external-ssl-comparisons}

\input{sec/5_conclusion}

\clearpage

{
    \small
    \bibliographystyle{ieeenat_fullname}
    \bibliography{main}
}

\clearpage
\clearpage


\setcounter{figure}{0}\renewcommand{\thefigure}{S\arabic{figure}}
\setcounter{table}{0}\renewcommand{\thetable}{S\arabic{table}}
\setcounter{equation}{0}\renewcommand{\theequation}{S\arabic{equation}}

\end{document}

%% file: sec/0_abstract.tex
\begin{abstract}
Self-supervised learning (SSL) methods have become a dominant paradigm for creating general purpose models whose capabilities can be transferred to downstream supervised learning tasks. However, most such methods rely on vast amounts of pretraining data. This work introduces Subimage Overlap Prediction, a novel self-supervised pretraining task to aid semantic segmentation in remote sensing imagery that uses significantly lesser pretraining imagery. Given an image, a sub-image is extracted and the model is trained to produce a semantic mask of the location of the extracted sub-image within the original image. We demonstrate that pretraining with this task results in significantly faster convergence, and equal or better performance (measured via mIoU) on downstream segmentation. This gap in convergence and performance widens when labeled training data is reduced. We show this across multiple architecture types, and with multiple downstream datasets. We also show that our method matches or exceeds performance  while requiring significantly lesser pretraining data relative to other SSL methods. Code and model weights are provided at \href{https://github.com/sharmalakshay93/subimage-overlap-prediction}{github.com/sharmalakshay93/subimage-overlap-prediction}.
\end{abstract}

%% file: sec/1_intro.tex
\section{Introduction}
\label{sec:intro}

\input{figures/pretrain_flow}

Accurate and timely Land Cover Classification (LCC) derived from remote sensing (RS) imagery is a foundational requirement for understanding and managing global environmental processes. The resultant geospatial data drives critical applications across numerous sectors, including monitoring large-scale land surface changes (such as urbanization and deforestation), informing efforts for detecting biodiversity loss, enhancing disaster prevention strategies (e.g., flood and wildfire risk modeling), and optimizing agriculture success tracking through precision farming and crop yield estimation. While the proliferation of remote sensing data provides an unprecedented opportunity to address these challenges, the effective deployment of state-of-the-art deep learning (DL) models is fundamentally hindered by two primary constraints: the inherent necessity for large, diverse training datasets and the exorbitant cost associated with generating high-quality, pixel-level ground truth labels. Addressing this annotation bottleneck is paramount for achieving scalable, operational LCC systems; natural directions of research include advanced machine learning paradigms, such as weak supervision, semi-supervised learning, and the development of more effective feature representation techniques, to unlock the full potential of remote sensing data for global monitoring.

Within the advanced machine learning paradigms necessary to overcome the labeling constraint, semi-supervised learning (SSL) methodologies can be broadly categorized based on their underlying mechanism for leveraging unlabeled data. One category of methods involves generative approaches, where the model learns effective feature representations by generating original images or reconstructing input data. A second group consists of discriminative methods, which involve training models using auxiliary - or pretext - tasks, such as predicting the relative position between image patches or enforcing consistency regularization under different data perturbations. A third category consists of contrastive methods, which extracts features by maximizing the similarity (or minimizing the distance) between the latent representations of positive instances (e.g., augmented views of the same image) and minimizing the similarity (or maximizing the distance) between representations of unrelated samples.

A key challenge inherent in modern SSL techniques, particularly contrastive and generative methods, is their reliance on massive amounts of unlabeled data and high computational resources. Typical state-of-the-art SSL approaches operate in a task-agnostic pre-training paradigm, making use of very large datasets to produce general-purpose foundation models that generalize well to multiple downstream tasks and datasets. However, there is significant value in exploring alternative techniques which have a narrower scope, specifically focusing on achieving good downstream results using only a limited amount of unlabeled data and minimal compute power. Such resource-efficient, task-aligned pretraining methods focus on solving specific problems, such as (in the remote sensing domain) specialized crop type identification in small regions or rapid localized disaster assessment, allowing researchers to make headway in those specific areas quickly and cheaply. This reduces the barrier to utilizing powerful deep learning models for researchers with limited infrastructural support.

This work addresses the need for computationally inexpensive and data-efficient feature learning by exploring task-aware self-supervision using a novel discriminative self-supervised spatial auxiliary task. This task, which we refer to as Subimage Overlap prediction, predicts the location of a subimage within the larger image from which it is selected, The task teaches the model to learn visual features that are highly transferable to downstream tasks. We demonstrate the effectiveness of this approach for remote sensing problems, specifically Land Cover Classification (LCC), where the learned features capture essential spatial and contextual information about ground objects. An overview of Subimage Overlap prediction is provided in Figure~\ref{fig:pretrain_flow}.

Our main contributions are threefold:
\begin{itemize}
\item The development and implementation of a resource-efficient Subimage Overlap prediction auxiliary task tailored specifically for remote sensing imagery, enabling the model to learn meaningful spatial and contextual feature representations without requiring labeled data.

\item A series of experiments validating the effectiveness of the task-aware pre-training using Subimage Overlap prediction, by comparing the performance of the learned features against baselines pretrained on datasets like ImageNet and LVD-142M.

\item A comprehensive analysis of the transfer learning capabilities of our method for semantic segmentation in remote sensing imagery by varying the downstream data distribution, and comparing performance against other competitive/state-of-the-art SSL approaches.
\end{itemize}

The remainder of this paper is structured as follows. Section~\ref{sec:related_work} discusses related work in semi-supervised learning for remote sensing and further motivates the proposed approach. Section~\ref{sec:subimage_overlap_pretraining} contains a detailed description of the Subimage Overlap pre-training method, including architecture decisions, implementation details, and experimental results validating its feature learning capabilities. Section~\ref{sec:downstream} discusses the use of Subimage Overlap in downstream transfer learning settings and analyzes the experimental results compared to well-known competitive/state-of-theart benchmarks. Finally, a summary of the contributions is provided in Section~\ref{sec:conclusion}.

%% file: figures/pretrain_flow.tex
\begin{figure*}[t]
  \centering
  \includegraphics[width=\linewidth]{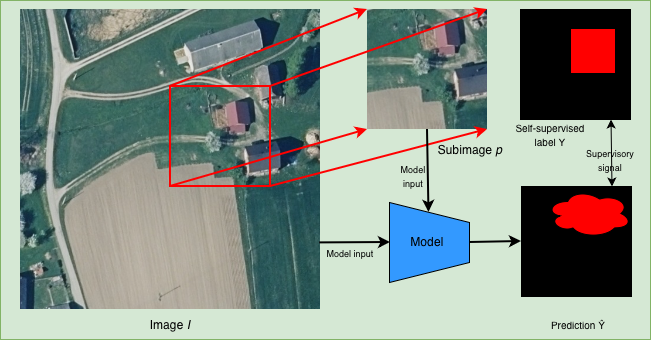}
  \caption{Overview of the Subimage Overlap pretraining process. The model receives an input image and a selected subimage, and learns to predict a binary mask indicating the subimage’s location within the image.}
  \label{fig:pretrain_flow}
\end{figure*}

%% file: sec/2_related_work.tex
\section{Related Work}
\label{sec:related_work}

Current state-of-the-art machine learning approaches require vast amounts of labeled data to be trained successfully. However, annotating sufficiently large amounts of data with high-quality labels for training such models is often prohibitively expensive, especially for challenging tasks involving real-world data like remote sensing. Various strategies have been used to mitigate the data annotation bottleneck. These strategies include: Unsupervised Learning, which directly estimates a model using only the raw data, without any supervisory signal (e.g., clustering or dimensionality reduction); Semi-Supervised Learning, which combines the use of small amounts of in-domain labeled training data with much larger amounts of unlabeled data; Weakly Supervised Learning, which relies on noisy, incomplete, or coarse-grained labels (e.g., using image-level tags for a pixel-level segmentation task); and Meta-Learning, which are algorithms designed to learn how to learn—that is, to rapidly adapt to a new task using only minimal data. More recently, Self-Supervised Learning (SSL) techniques have been widely employed; in SSL, the goal is to learn a model using only 'natural' supervision, i.e. a supervisory signal derived directly from the data itself, following the intuition that such a model can extract features that are highly transferable and of use to a broad variety of other downstream tasks \cite{Liu2023Self}. The resulting pre-trained models serve as excellent feature extractors, significantly reducing the labeled data requirements for subsequent transfer learning steps.

SSL techniques are typically categorized based on their underlying pretext task. Generative methods learn powerful representations by training the model to reconstruct or generate the input data, thereby capturing the complete data distribution. Key examples include traditional Autoencoders (AEs) and their advanced variations like Sparse Autoencoders \cite{lu2017remote}, Denoising AEs \cite{ZHANG201624}, Variational Autoencoders (VAEs) \cite{Palsson2022Blind}, and more recently, the highly effective Masked Autoencoders (MAEs) \cite{He2022Masked}. Generative Adversarial Networks (GANs) have also been used in a self-supervised context to learn meaningful representations \cite{Zhu2018Generative, Jin2023Adversarial}. A second, highly influential class of methods includes Predictive (or Discriminative) methods, where a suitable pretext task is selected for which labels can be generated directly from the data, and a model is trained to predict such labels. These methods exploit the inherent structure of the data, such as spatial or temporal context, to enforce learning. Predictive methods are diverse. Spatial tasks are most common, including tasks such as predicting the relative position~\cite{doersch2015unsupervised} or co-occurence\cite{isola2015learning} of random patch pairs, predicting the rotation angle applied to an image (rotation prediction) \cite{Wen2021Rotation}, recovering a missing patch of the input (inpainting) \cite{singh2018self}, or predicting the correct order of a shuffled grid of patches (Jigsaw Puzzle). Other predictive tasks involve spectral methods (e.g., predicting the colors in an image based on its grayscale version, \cite{Zhang2016Colorful, Vincenzi2020Color}), temporal tasks (e.g., predicting the order of frame sequences in a video \cite{Yuan2021Self}), and miscellaneous tasks like counting visual primitives, spotting artifacts \cite{He2022Self}, or correlating visual and geographical information \cite{Li2021Geographical}. The success of a predictive method heavily relies on the design of a pretext task that is challenging enough to necessitate the learning of useful, high-level features for the downstream application.

The third, and currently dominant, category of SSL methods is Contrastive Self-Supervision. The fundamental principle of contrastive learning is to learn feature representations by maximizing the similarity between different augmented views of the same image (positive pairs) and minimizing the similarity between views of different images (negative pairs) \cite{Liu2023Self}. This approach assumes that semantically similar content should be close in the embedding space. Contrastive methods can be further subdivided based on how they generate or manage these pairs: Negative Sampling Methods directly use a loss function (like Triplet Loss or the InfoNCE Loss) to explicitly push apart negative samples. Highly successful algorithms in this area include SimCLR \cite{Chen2020Simple}, which relies on large batch sizes, and MoCo \cite{He2020Momentum}, which uses a memory bank or a momentum encoder to manage a large queue of negative samples. Remote sensing applications of SimCLR include \cite{Zhao2022Hyperspectral} and \cite{Zhu2022SCEADNet}. Clustering Methods integrate instance-discrimination with cluster assignment to learn representations, with examples such as DeepCluster \cite{Caron2018Deep}, LocalAgg, and PCL. DeepCluster was used for both change detection \cite{Saha2022Self} and semantic segmentation \cite{Saha2022Unsupervised}. In contrast, knowledge distillation methods abandon the need for explicit negative pairs entirely. These methods, like BYOL \cite{Grill2020Bootstrap} and SimSiam \cite{Chen2021Exploring}, often employ two interacting networks (an online network and a target network, typically updated via a momentum strategy) that predict the representation of one view of an image from another view, without collapsing to a trivial solution. A notable and highly effective example is DINO (Data-efficient Image Transformers) \cite{Caron2021Emerging}. DINO is a self-distillation approach that leverages Vision Transformers (ViT) to learn image representations by aligning the output of a student network with the output of a momentum-updated teacher network, using a centering and sharpening mechanism to prevent collapse. DINO is particularly renowned for its ability to learn high-quality, dense features that reveal clear, semantic segmentation properties without requiring any explicit labels, making it a powerful foundation for vision tasks. Finally, Redundancy Reduction Methods, such as Barlow Twins \cite{Zbontar2021Barlow} and VICReg \cite{Bardes2022VICReg}, learn representations by making the cross-correlation matrix between the outputs of two augmented views of the same sample close to the identity matrix, effectively enforcing that the learned features are non-redundant.

Building upon the initial success of DINO, follow-up work has focused on improving scalability and robustness. DINOv2 \cite{oquab2023dinov2}, for instance, represents a significant advancement by scaling the training to billions of images, leveraging a curated, large-scale dataset, and integrating technical improvements such as specialized data processing and training stability enhancements. This massive scale-up resulted in models that achieve state-of-the-art performance across a wide range of computer vision benchmarks without requiring fine-tuning for many tasks. Related distillation approaches, such as i-Jepa \cite{assran2023self}, move towards non-generative, predictive modeling by predicting masked-out image content in the latent representation space, rather than the pixel space. This focus on learning meaningful semantic structures in the latent domain, as opposed to pixel-perfect reconstruction, continues the trend of making self-supervised models more powerful, versatile, and suitable for deployment as general-purpose foundation models. The sheer size and diversity of the unlabeled dataset used to train DINOv2 mean that the resulting ViT backbone is highly robust, capturing a broad and generalized set of visual features. This makes the DINOv2 backbone an ideal starting point for adapting to specific domains, such as remote sensing, as it provides a powerful, pre-trained feature extractor that minimizes the domain-specific pre-training effort required for subsequent self-supervision or fine-tuning approaches.

%% file: sec/3_subimage_overlap_pretraining.tex
\section{Subimage Overlap Pretraining}
\label{sec:subimage_overlap_pretraining}

Given an input image $\mathbf{I}$ of length $l$ and width $w$, 
a subimage $\mathbf{p}$ of length $p_l$ and width $p_w$ is selected such that 
$p_l \leq l$ and $p_w \leq w$. 

Let $\mathbf{Y}$ denote the ground truth semantic mask and 
$\hat{\mathbf{Y}}$ denote the mask predicted by the model $M$. 
Both $\mathbf{Y}$ and $\hat{\mathbf{Y}}$ share the same spatial dimensions as the input image 
$\mathbf{I}$, and $\mathbf{Y}$ contains positive labels at pixel locations corresponding to the 
selected subimage $\mathbf{p}$ and zeros elsewhere.

The model is given input $\mathbf{X}$ which is a combination of the full image $\mathbf{I}$ and the selected subimage $\mathbf{p}$, 
and is trained to predict $\hat{\mathbf{Y}}$ 
where positive pixels indicate the location of the selected subimage.

\begin{align}
    \hat{\mathbf{Y}} &= M(\mathbf{X}), \nonumber \\
    \mathbf{Y}_{i,j} &=
    \begin{cases}
        1, & \text{if } (i,j) \in \mathbf{p}, \\
        0, & \text{otherwise}.
    \end{cases} \\
    \mathbf{Y},\, \hat{\mathbf{Y}} &\in \{0,1\}^{l \times w}, \nonumber \\
    \label{eq:subimage-prediction}
\end{align}

The goal of this pretraining task is to have the model learn visual features 
that are transferable to downstream tasks on remote sensing imagery. 
Since localizing a subimage within a larger image requires identifying correspondences 
between the subimage and the full image—leveraging both low-level cues (e.g., edges, colors, textures) 
and high-level cues (e.g., shapes, objects, spatial context)—we hypothesize that this task 
encourages the model to learn semantically meaningful representations.

Because the labels for this task are derived directly from the image itself, 
this pretraining objective is fully self-supervised and requires no human annotation.

\input{figures/subimage_examples}

\subsection{Architecture}

\subsubsection{DinoV2 backbone}
\label{subsubsec:dinov2-arch}

A {DINOv2 ViT-S/14}~\cite{oquab2023dinov2} model is adapted for this task. 
To enable multi-image input, the token sequence of the full image $\mathbf{I}$ 
is concatenated with that of the subimage $\mathbf{p}$, and a trainable separator token 
$\langle\mathrm{SEP}\rangle$ is inserted between them:
\[
\mathbf{X}
= \big[\mathrm{Enc}(\mathbf{I});\; \langle\mathrm{SEP}\rangle;\; \mathrm{Enc}(\mathbf{p})\,\big],
\]
where $\mathrm{Enc}(\cdot)$ denotes the ViT patch embedding and positional encoding, and $\langle\mathrm{SEP}\rangle$ is a learnable parameter 
optimized jointly with the model. The ViT class token $\mathrm{[CLS]}$ is dropped. In downstream finetuning tasks, the $\langle\mathrm{SEP}\rangle$ token is not used.

A lightweight convolutional decode head that maps the ViT patch-level features to a dense semantic mask is appended to the DINOv2 encoder. Patch embeddings are reshaped into a 2D feature map, passed through a small stack of convolutional layers, and then upsampled to the original image resolution to produce per-pixel class predictions. We intentionally use this simple decode head to (i) test whether the task is learnable with a commonly used backbone but with minimal added architectural complexity and (ii) assess whether this pretraining task yields an encoder whose learned representations 
are semantically meaningful, independent of decoder complexity. 

\subsubsection{ResNet-50 Backbone}
\label{subsubsec:resent50-arch}

Additionally, we train a ResNet-50~\cite{he2016deep} backbone for 
some experiments. 

In this setup, a dual-encoder network architecture is used where one encoder takes as input the full image $\mathbf{I}$ and the other takes as input the subimage $\mathbf{p}$. These produce feature maps $\mathbf{F}_I$ and $\mathbf{F}_p$ for their respective inputs. The subimage features 
$\mathbf{F}_p$ are first bilinearly upsampled to match the dimensions of 
$\mathbf{F}_I$ and then concatenated along the channel dimension, thus yielding a combined 
representation. This concatenated 
feature map is passed through a fusion module (a small convolutional block) that mixes and 
reduces the channels to a shared representation, which is then fed into a decode head to 
predict the segmentation mask for the Subimage Overlap prediction task.

For downstream fine-tuning, we load and use only the full-image encoder

\subsection{Training Subimage Overlap Prediction}
\label{subsec:train_pos_details}

To ensure the feasibility of the task-aware pretraining method using Subimage Overlap, an initial set of experiments was performed using the LandCoverAI~\cite{boguszewski2021landcover} dataset. Viable training parameters / hyper-parameters were evaluated by training the Subimage Overlap segmentation task using a DINOv2 ViT-S/14 backbone and decoder head as described in Section~\ref{subsubsec:dinov2-arch}.

Each original $512 \times 512$ image was resized to $224 \times 224$. Training / validation / test splits specified in~\cite{boguszewski2021landcover} were used resulting in 7,470 training and 1,602 validation images. The test set was not used to prevent data leakage in downstream landcover segmentation on this dataset. 

The following variations were explored to identify the optimal hyperparameters for the pretraining task. 

\begin{enumerate}
    \item \textbf{Loss:} Binary Cross-Entropy vs.\ Focal Loss~\cite{lin2017focal}.
    \item \textbf{Augmentations:}
    \begin{enumerate}
        \item \textbf{Position-based:} vertical and horizontal flips.
        \item \textbf{Color-based:} brightness, contrast, saturation, and hue jitter.
        \end{enumerate}
    \item \textbf{Subimage size:} $56 \times 56$ pixels, $112 \times 112$ pixels.
\end{enumerate}

Augmentations are first applied to the full image prior to subimage selection, and then independently applied to the selected subimage. This aims to improve robustness by exposing the model to cases where the subimage is a mirrored or color-perturbed version of the full image.

We use an initial learning rate of $1\times 10^{-4}$ with the AdamW optimizer~\cite{loshchilov2019adamw} 
and a cosine annealing learning rate schedule~\cite{loshchilov2017sgdr}. 
For the focal loss, we set $\gamma = 1.5$ and $\alpha = [0.25,\, 0.75]$ to address the imbalance 
between background and subimage pixels. Training is performed for 150 epochs with a batch size of 64 
on a single NVIDIA~T4 GPU.

\input{figures/subimage_overlap_epochs}

\subsection{Evaluation / Results - Subimage Overlap Prediction}
\label{subsec:pos_pretrain_eval}

Since this is a semantic segmentation task with imbalanced classes, we evaluate it using mean intersection-over-union (mIoU).

\begin{enumerate}
    \item \textbf{Loss:} Using Focal Loss results in better performance, which is consistent with its effectiveness in handling class imbalance in the occurrence of positive and negative pixels.

    \item \textbf{Position-based augmentations:} Applying spatial augmentations such as vertical and horizontal flips did not significantly change performance.
    
    \item \textbf{Color-based augmentations:} These augmentations degrade performance and introduce training instability, observed as large fluctuations in validation accuracy (despite the fact that augmentations are only applied to train samples). We hypothesize that this occurs because color and edge information are critical for establishing correspondences between the subimage $\mathbf{p}$ and the full image $\mathbf{I}$. Another observation was that train performance lags behind validation due to color jitter augmentation. As is conventional, jittering is applied only on the train split. 
    
    \item \textbf{Subimage size:} Performance decreases when the subimage size is too small, likely because smaller subimages contain insufficient semantic context for reliable overlap localization.
\end{enumerate}

\input{tables/subimageoverlap-pretrain-landcover}

Table~\ref{tab:subimageoverlap-pretrain-landcover} summarizes the pretraining results. Pretraining with no augmentations performs the best, although by only a small margin compared to pretraining with random flips. Both of these performed significantly better than any training with jittering included. Figure~\ref{fig:subimage_examples} shows some predictions from the \textit{With flip} model, and Figure~\ref{fig:subimage_overlap_epochs} shows how model performance evolves over training epochs.

As upstream performance is not always predictive of downstream performance, we finetune and evaluate each model in Table~\ref{tab:subimageoverlap-pretrain-landcover} on the downstream segmentation task to assess both (i) the benefit of pretraining over no pretraining and (ii) whether upstream rankings between models are preserved downstream.

Note that all models weights initialization and standard dataset splits are used via 
HuggingFace\footnote{\url{https://huggingface.co/}} or 
TorchGeo~\cite{stewart2025torchgeo}. \textit{No pretraining} variants are initialized with ImageNet weights for ResNet-50 based models, and LVD-142M~\cite{oquab2023dinov2} weights for DinoV2 based models. Pretrained variants are initialized with the same before undergoing pretraining.

%% file: figures/subimage_examples.tex
\begin{figure*}[t]
  \centering
  \begin{subfigure}{0.24\linewidth}
    \includegraphics[width=\linewidth]{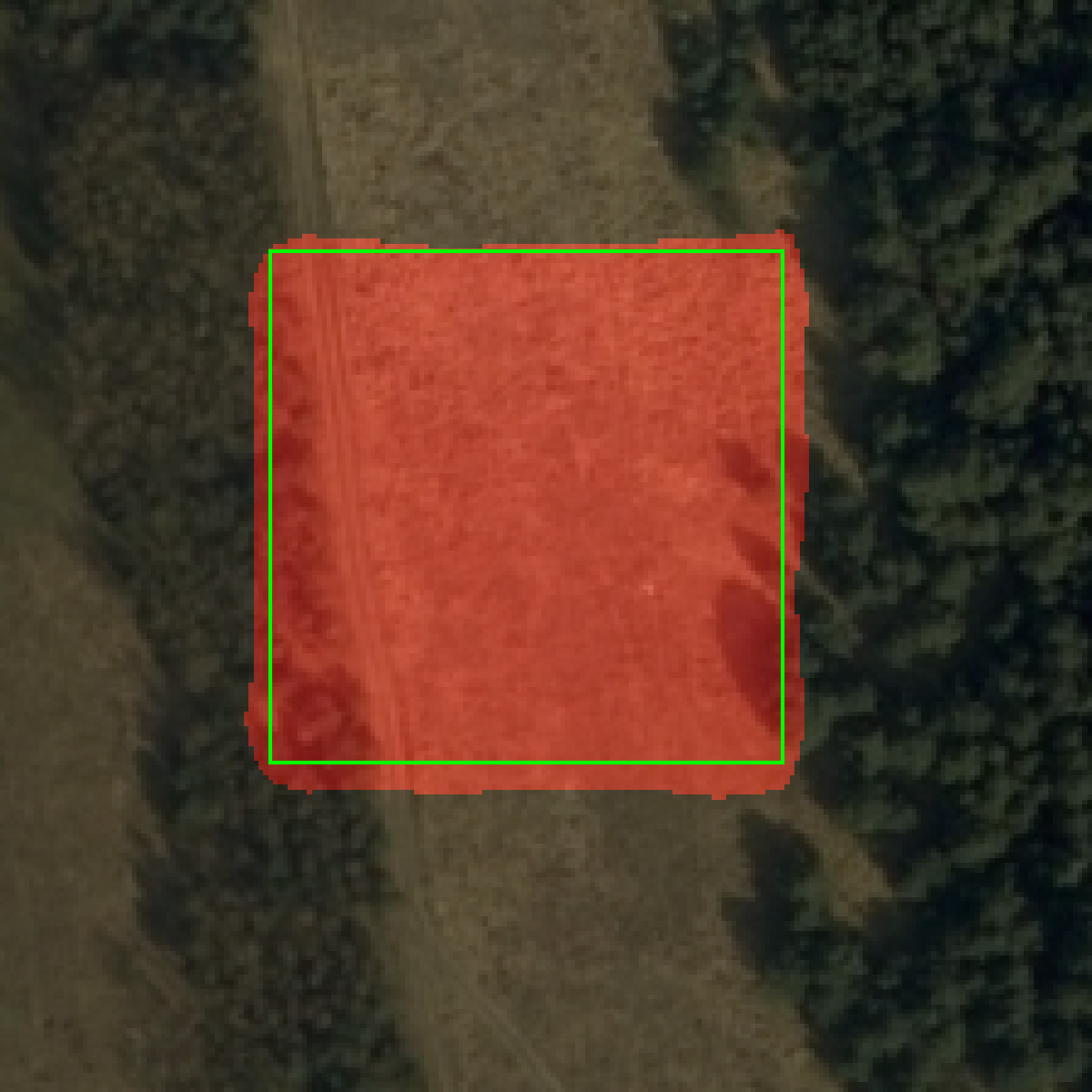}
    \label{fig:seg-lcai}
  \end{subfigure}\hfill
  \begin{subfigure}{0.24\linewidth}
    \includegraphics[width=\linewidth]{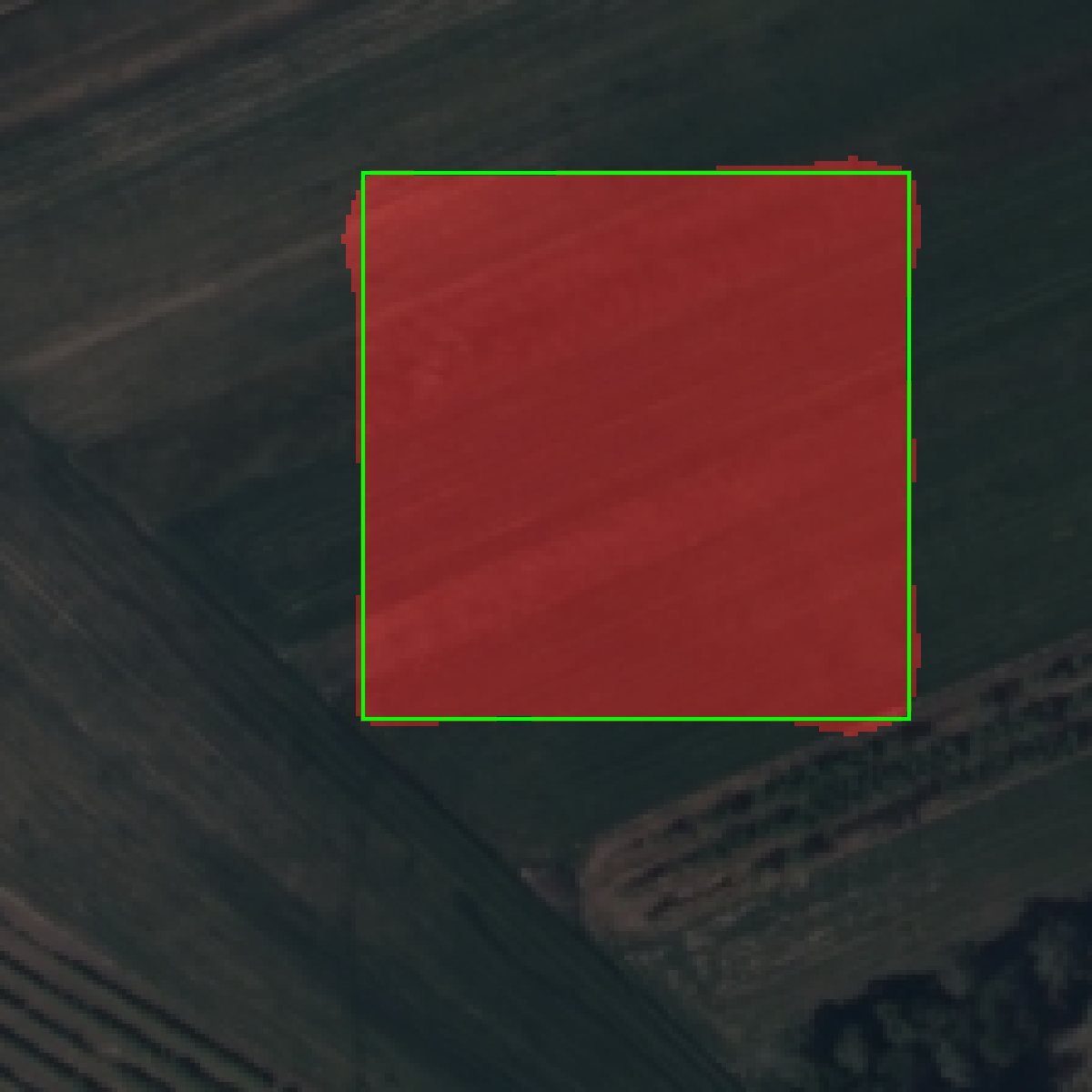}
    \label{fig:seg-deepglobe}
  \end{subfigure}\hfill
  \begin{subfigure}{0.24\linewidth}
    \includegraphics[width=\linewidth]{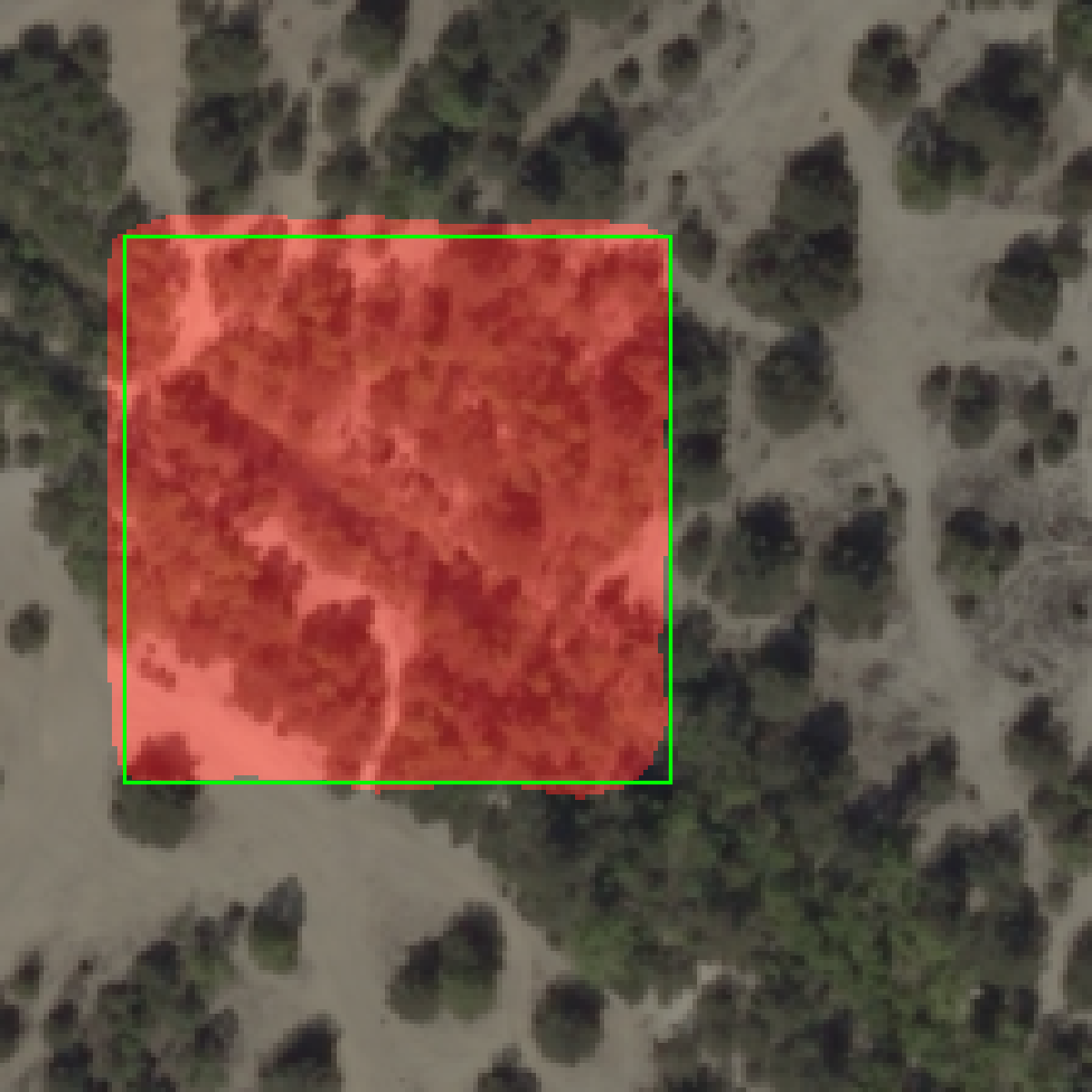}
    \label{fig:seg-loveda}
  \end{subfigure}\hfill
  \begin{subfigure}{0.24\linewidth}
    \includegraphics[width=\linewidth]{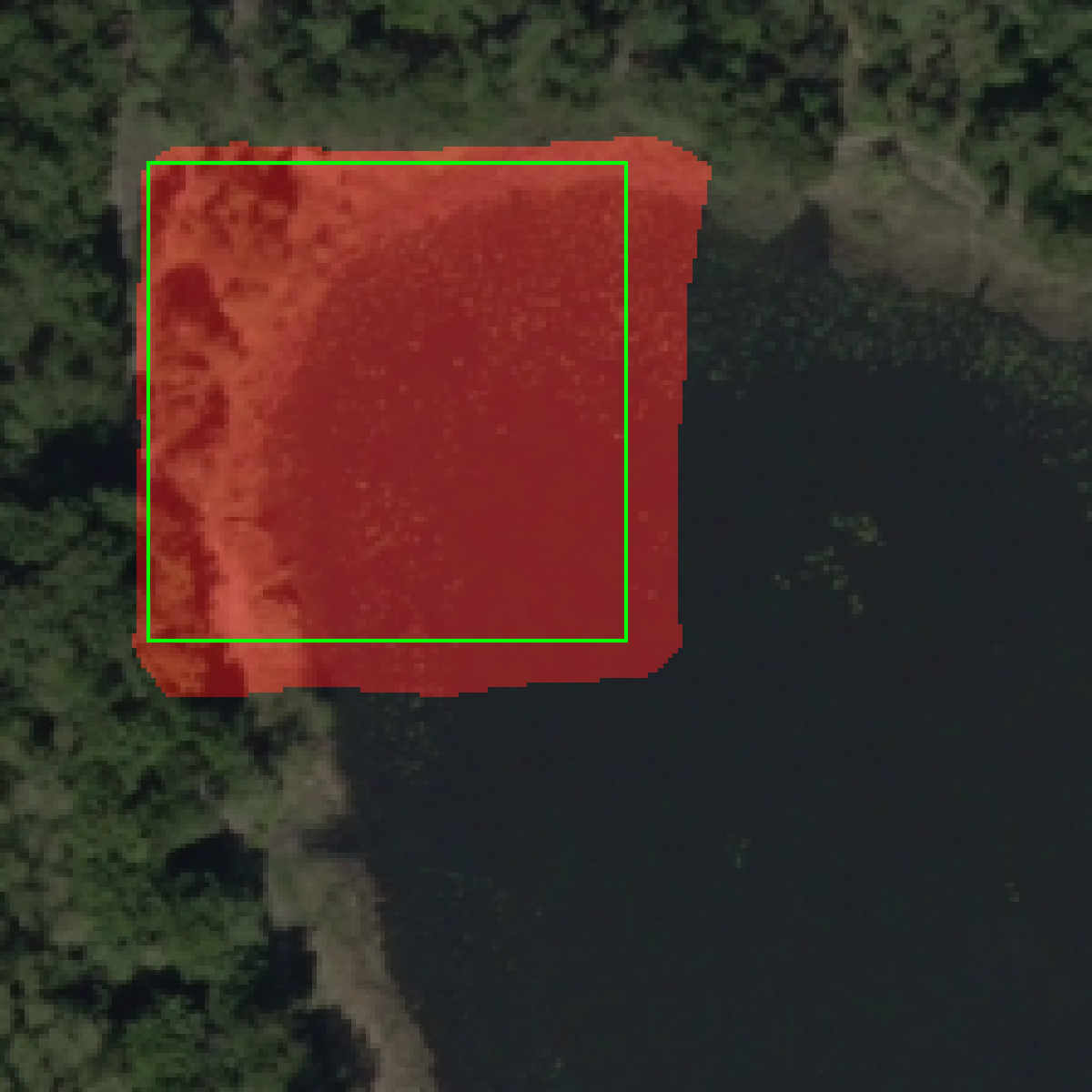}
    \label{fig:seg-oem}
  \end{subfigure}
  \caption{\small Subimage Overlap prediction examples. The green square represents the selected subimage / ground truth; the red mask shows the predictions by a DinoV2 backbone model.}
  \label{fig:subimage_examples}
\end{figure*}

%% file: figures/subimage_overlap_epochs.tex
\begin{figure*}[t]
  \centering
  \begin{subfigure}[t]{0.36\textwidth}
    \centering
    \includegraphics[width=\linewidth]{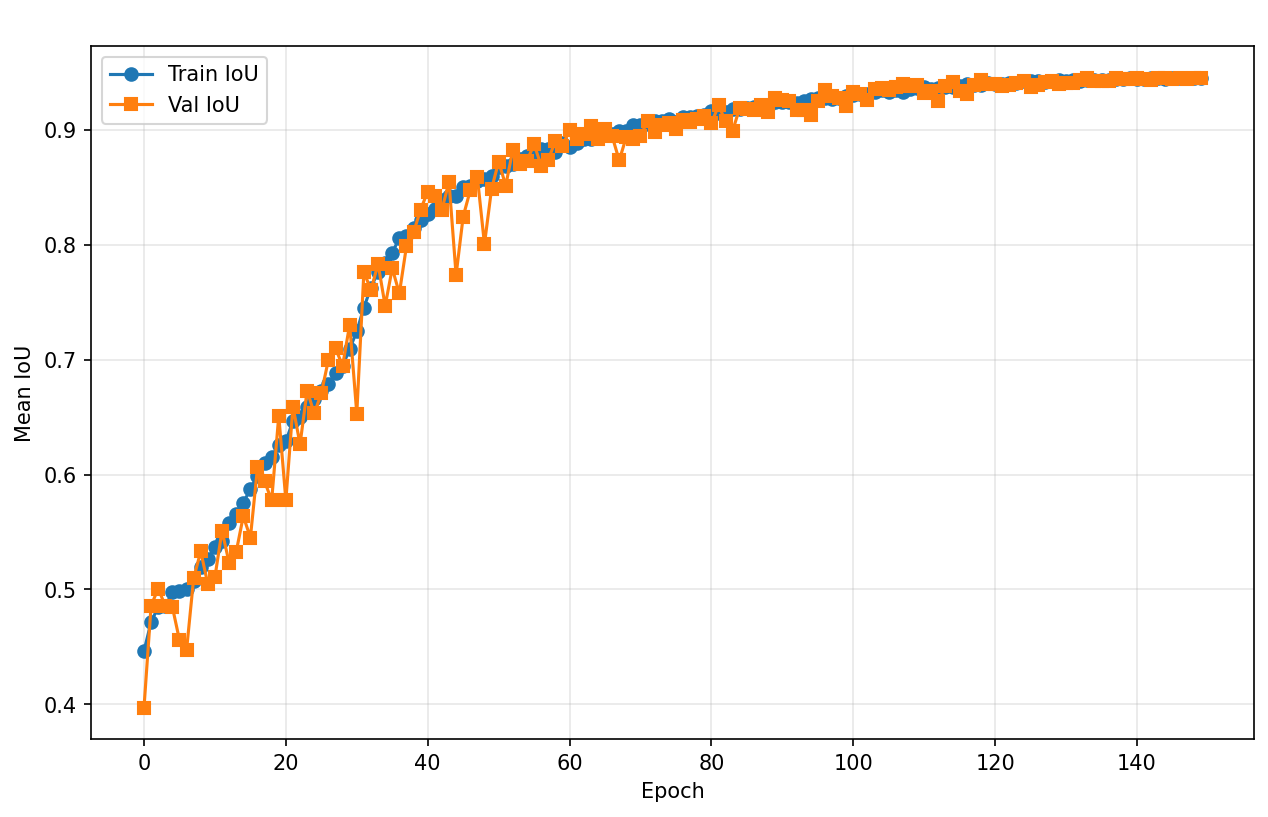}
    \caption{Train-val mIoU by epoch}
    \label{fig:panel-1}
  \end{subfigure}\hfill
  \begin{subfigure}[t]{0.20\textwidth}
    \centering
    \includegraphics[width=\linewidth]{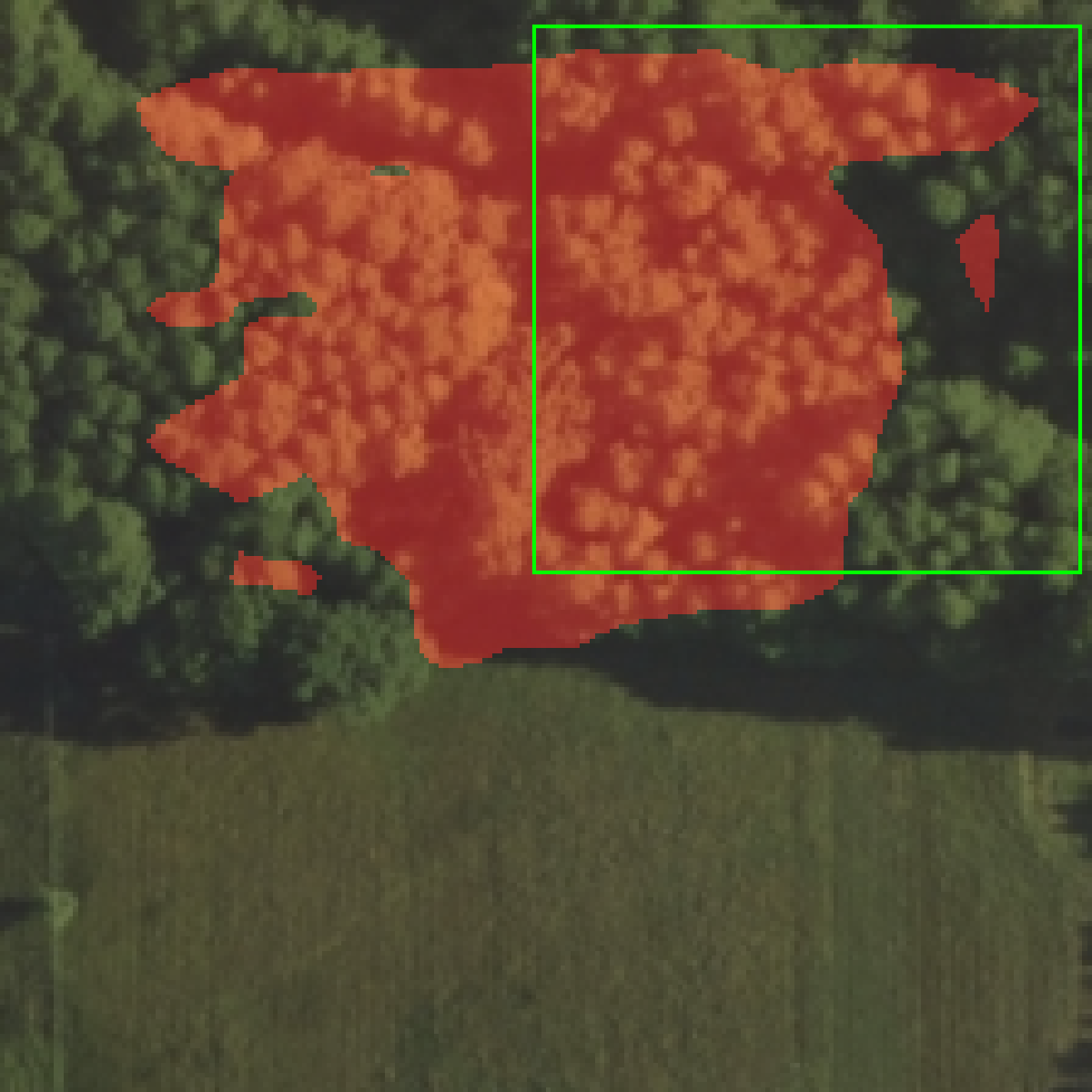}
    \caption{Epoch 30 prediction}
    \label{fig:panel-2}
  \end{subfigure}\hfill
  \begin{subfigure}[t]{0.20\textwidth}
    \centering
    \includegraphics[width=\linewidth]{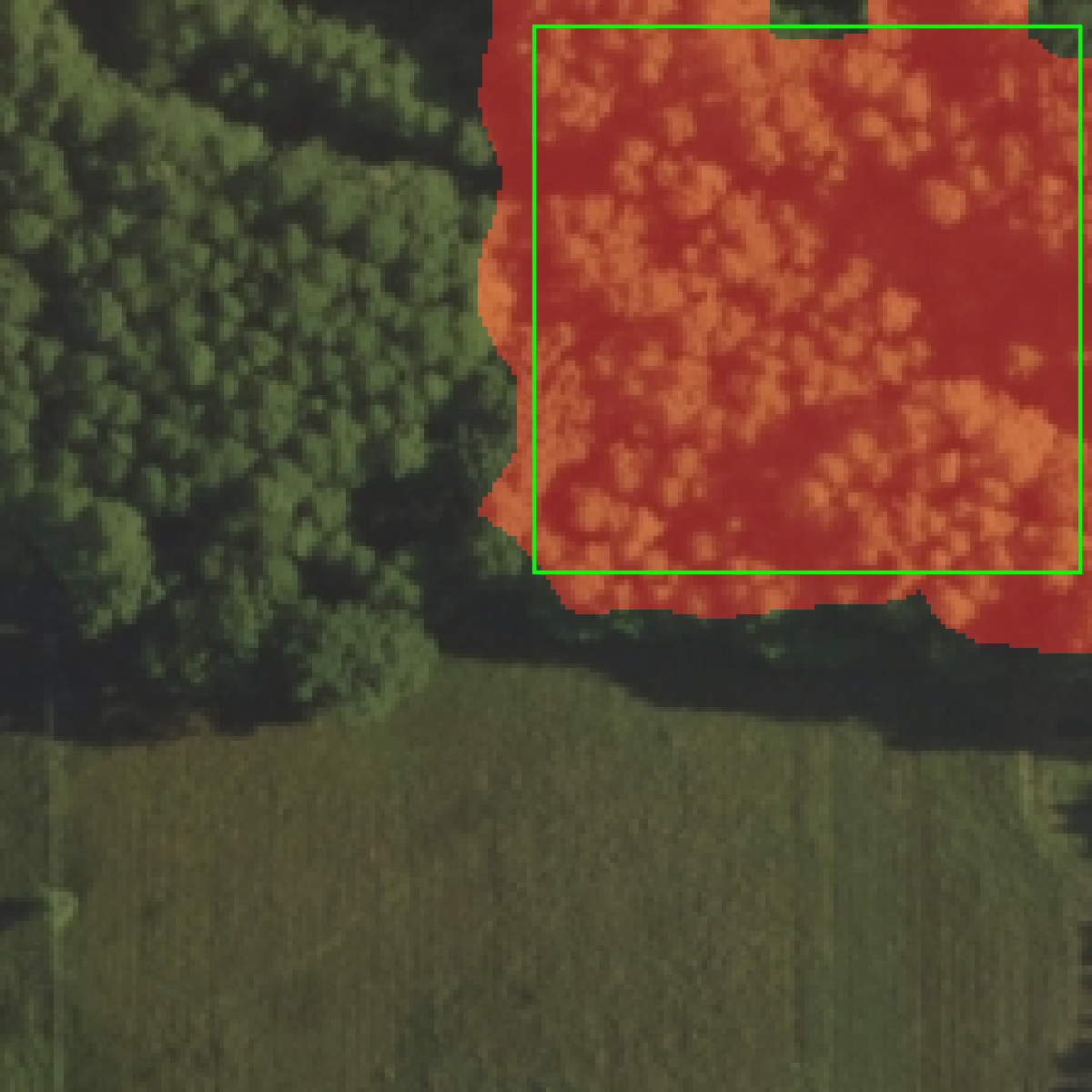}
    \caption{Epoch 60 prediction}
    \label{fig:panel-3}
  \end{subfigure}\hfill
  \begin{subfigure}[t]{0.20\textwidth}
    \centering
    \includegraphics[width=\linewidth]{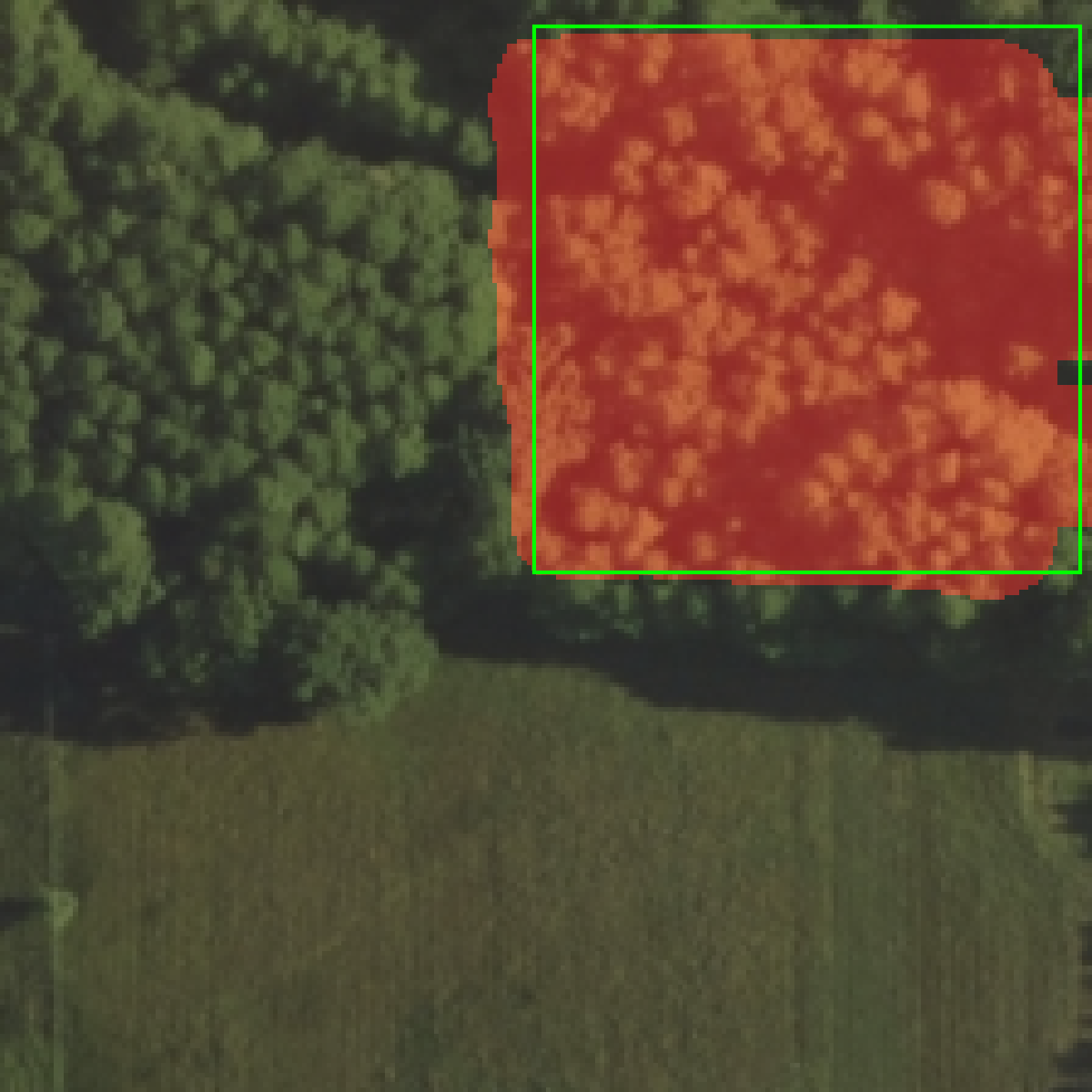}
    \caption{Epoch 150 prediction}
    \label{fig:panel-4}
  \end{subfigure}

  \caption{\small Subimage Overlap pretraining with LandCoverAI: train-val mIoU curves and predictions by epoch. Green boxes represent the selected subimage / ground truth; red mask represents the prediction after $k$ training epochs.}
  \label{fig:subimage_overlap_epochs}
\end{figure*}

%% file: tables/subimageoverlap-pretrain-landcover.tex
\begin{table}[t]
  \centering
  \setlength{\tabcolsep}{3pt}
  \scriptsize
  \begin{tabular}{@{}lc@{}}
    \toprule
    Model & Val IoU (pretraining) \\
    \midrule
    No augmentations   & 0.9605 \\
    w flip             & 0.9434 \\
    w jitter           & 0.8052 \\
    w flip + jitter    & 0.7159 \\
    \bottomrule
  \end{tabular}
  \caption{\small Validation IoU for subimage-overlap pretraining  on LandCoverAI data (higher is better).}
  \label{tab:subimageoverlap-pretrain-landcover}
\end{table}

%% file: sec/4_downstream.tex
\section{Downstream transfer learning}
\label{sec:downstream}

We restrict downstream evaluation to remote sensing imagery and use RGB channels exclusively for semantic segmentation. Unless stated otherwise, we finetune all layers and report mean IoU (mIoU). 

%% file: sec/4_1_po-downstream-ablation.tex
\subsection{Ablating the Task-Aware Pretraining for Land-Cover Segmentation}
\label{subsec:po-downstream-ablation}

We transfer the 
upstream checkpoints from Section~\ref{subsec:pos_pretrain_eval} to a land-cover segmentation model and evaluate downstream
mIoU on the LandCoverAI dataset. To quantify convergence, we
report (i) the first epoch whose mIoU is within 10\% of that run’s own best mIoU
(relative; i.e., $\ge 0.9\times$ best) and (ii) the first epoch whose mIoU is
within 10 absolute percentage points of that best (absolute; i.e., $\ge$
best$-0.10$). We also report fixed-epoch snapshots (5/15/30/60/100) to illustrate
learning speed, applying a short moving-average smoothing to mitigate the effect
of outlier epochs. Finally, we assess label efficiency by repeating training with 50\% and 25\% of the labeled data.

Unless noted, we reuse the hyperparameters from Section~\ref{subsec:train_pos_details}.
Epochs are set to 100 and batch size to 128.
The focal-loss class weights \(\boldsymbol{\alpha}\) are set to the inverse square root of each class’s pixel-frequency in the training set. The focusing parameter \(\boldsymbol{\gamma}=1.5\)


\subsubsection{Evaluation / results}
\label{subsubsec:sem_seg_results}

\input{figures/seg-convergence-data-var}

\input{tables/seg-val-iou-summary}


As shown in Table~\ref{tab:seg-val-iou-summary}, the no-pretraining (LVD-142M weights) baseline converges slower and ultimately trails all pretrained variants, despite approaching their peak mIoU. Measuring the first epoch within 10\% of a model’s best validation IoU, nearly all pretrained runs hit the threshold well before the baseline; the same holds under a 10-percentage-point margin. Overall, pretraining strongly accelerates convergence and slightly boosts peak performance when all labeled training data is used.

Table~\ref{tab:seg-val-iou-convergence} summarizes how validation IoU evolves over the course 
of training, with all pretrained variants exhibiting faster convergence than the 
no-pretraining baseline. The full convergence curves for the \textit{No pretraining} and \textit{Pretrain w flip} variants are in ~\ref{fig:side-a}.

\input{tables/seg-val-iou-convergence}

The convergence and performance gaps widen as downstream training data is reduced,. Figure~\ref{fig:seg-convergence-data-var} shows IoU convergence and the best performance achieved 
with 100\%, 50\%, and 25\% of the labeled data. Because the upstream self-supervised task 
always leverages the full unlabeled dataset, these results indicate that the task-aware pretraining using Subimage Overlap is especially advantageous when unlabeled imagery is plentiful but labeled 
samples are scarce, a scenario that is common in remote sensing imagery.

%% file: figures/seg-convergence-data-var.tex
\begin{figure*}[t]
  \centering
  \begin{subfigure}{0.32\linewidth}
    \includegraphics[width=\linewidth]{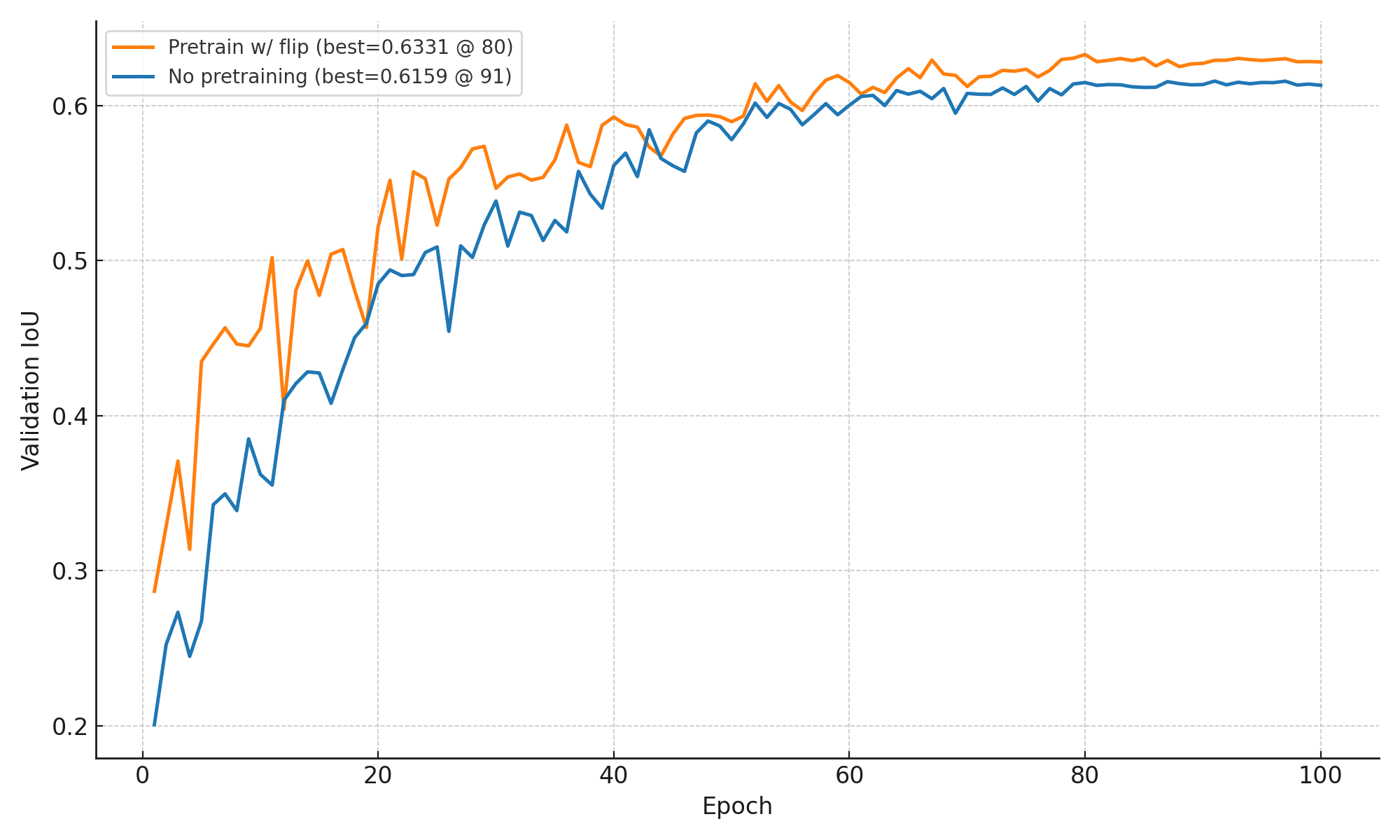}
    \caption{100\% labeled data}
    \label{fig:side-a}
  \end{subfigure}\hfill
  \begin{subfigure}{0.32\linewidth}
    \includegraphics[width=\linewidth]{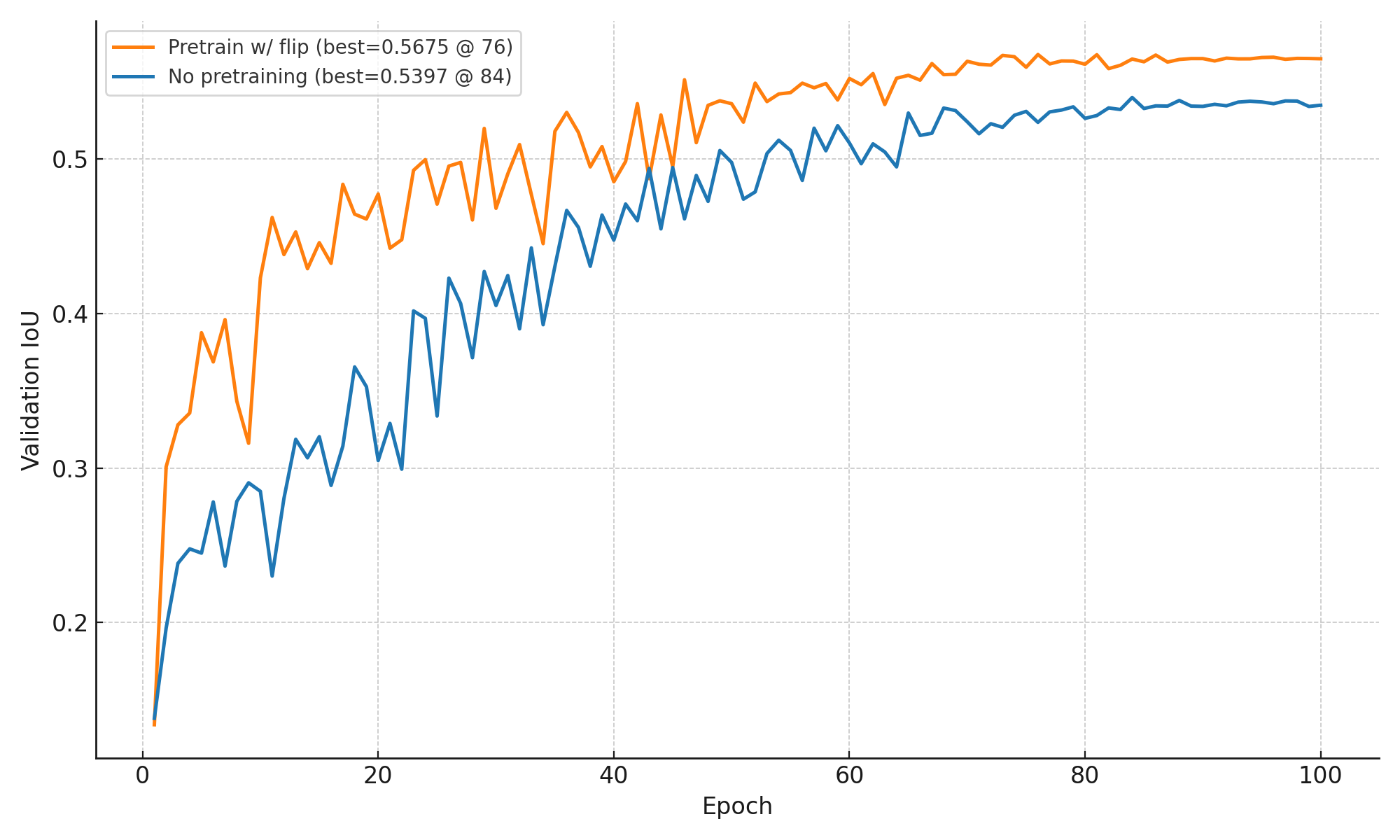}
    \caption{50\% labeled data}
    \label{fig:side-b}
  \end{subfigure}\hfill
  \begin{subfigure}{0.32\linewidth}
    \includegraphics[width=\linewidth]{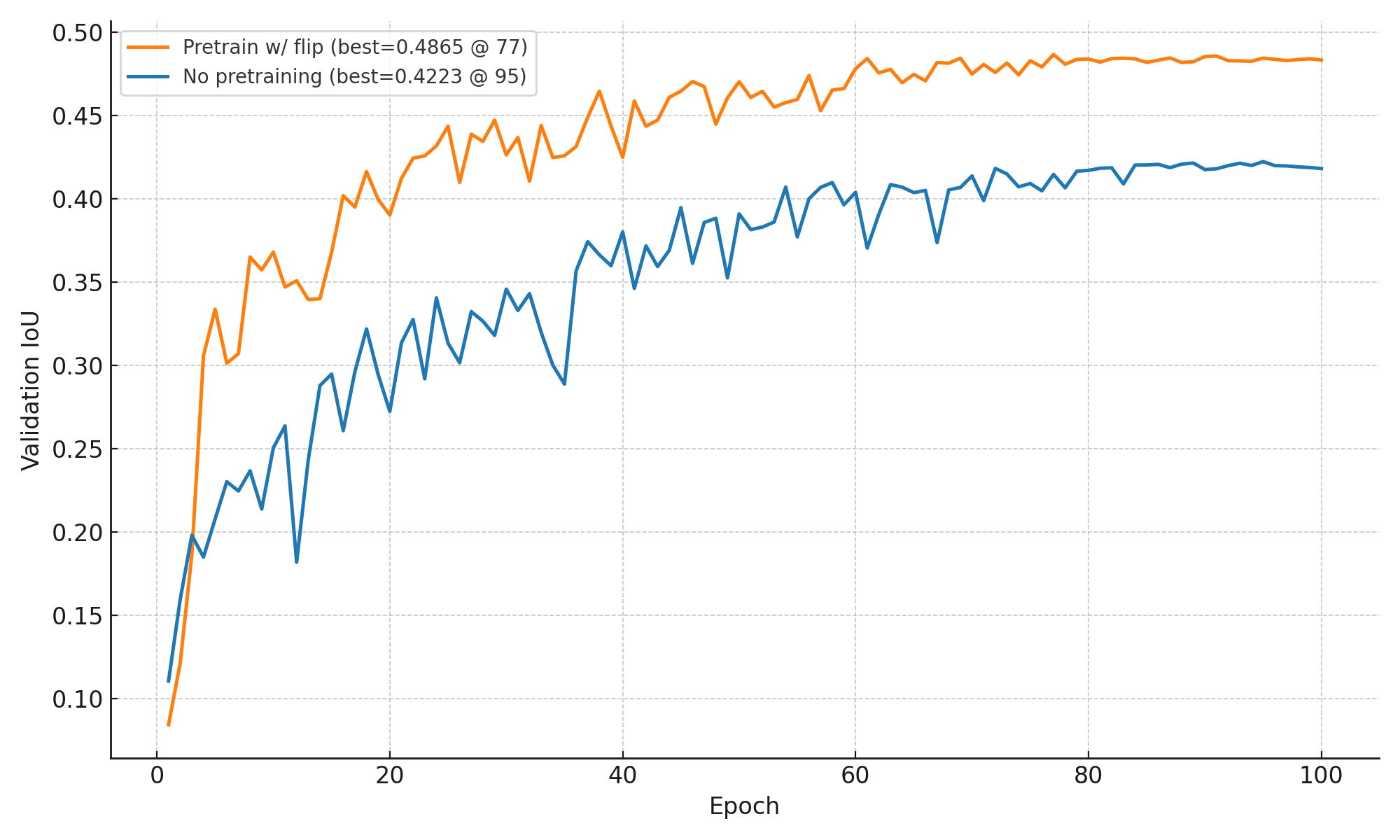}
    \caption{25\% labeled data}
    \label{fig:side-c}
  \end{subfigure}
  \caption{\small LandCoverAI segmentation: IoU convergence and best performance with varying amounts of labeled training samples.}
  \label{fig:seg-convergence-data-var}
\end{figure*}

%% file: tables/seg-val-iou-summary.tex
\begin{table}[t]
  \centering
  \setlength{\tabcolsep}{2pt}
  \scriptsize
  \begin{tabular}{@{}>{\raggedright\arraybackslash}p{0.17\columnwidth}
                  M{0.16\columnwidth}M{0.16\columnwidth}
                  M{0.20\columnwidth}M{0.12\columnwidth}@{}}
    \toprule
    \multicolumn{1}{c}{Model} & Best val IoU & Epochs within 10\% of best & Epochs within 10pp of best & Test IoU \\
    \midrule
    No pretraining        & \makecell{0.6159\\(Epoch 91)}                & 37 & 29 & .6265 \\
    \midrule
    Pretrain w/o augment  & \makecell{0.6324\\(Epoch 79)}                & 29 & 22 & -- \\
    \midrule
    Pretrain w flip       & \makecell{\textbf{0.6331}\\\textbf{(Epoch 80)}} & \textbf{28} & \textbf{21} & \textbf{.6355} \\
    \midrule
    Pretrain w jitter     & \makecell{0.6226\\(Epoch 93)}                & 33 & 23 & -- \\
    \midrule
    Pretrain w flip + jitter & \makecell{0.6247\\(Epoch 91)}             & 29 & 22 & -- \\
    \bottomrule
  \end{tabular}
  \caption{\small LandCoverAI segmentation: Best validation IoU per model (epoch shown in-cell), first epoch within 10\% of that best (relative), first epoch within 10 percentage points (absolute, best$-0.10$), and test IoU. Bold indicates the best in column.}
  \label{tab:seg-val-iou-summary}
\end{table}

%% file: tables/seg-val-iou-convergence.tex
\begin{table}[t]
  \centering
  \setlength{\tabcolsep}{2pt}
  \scriptsize
  \begin{tabular}{@{}>{\raggedright\arraybackslash}p{0.17\columnwidth}*{5}{M{0.145\columnwidth}}@{}}
    \toprule
    \multicolumn{1}{c}{Model} & \multicolumn{5}{c}{Val IoU at epoch (smoothed; raw values also shown)} \\
    \cmidrule(lr){2-6}
    \multicolumn{1}{c}{} & 5 & 15 & 30 & 60 & 100 \\
    \midrule
    No pretraining
      & \makecell{0.285\\raw: 0.267}
      & \makecell{0.421\\raw: 0.428}
      & \makecell{0.524\\raw: 0.539}
      & \makecell{0.600\\raw: 0.600}
      & \makecell{0.614\\raw: 0.613} \\
    \midrule
    Pretrain w/o augment
      & \makecell{0.392\\raw: 0.406\\(+0.108)}
      & \makecell{0.490\\raw: 0.464\\(+0.069)}
      & \makecell{0.559\\raw: 0.535\\\textbf{(+0.035)}}
      & \makecell{0.613\\raw: 0.611\\(+0.013)}
      & \makecell{0.626\\raw: 0.626\\(+0.013)} \\
    \midrule
    Pretrain w flip
      & \makecell{0.398\\raw: 0.435\\\textbf{(+0.113)}}
      & \makecell{0.494\\raw: 0.478\\\textbf{(+0.073)}}
      & \makecell{0.558\\raw: 0.547\\(+0.034)}
      & \makecell{0.614\\raw: 0.615\\\textbf{(+0.014)}}
      & \makecell{0.628\\raw: 0.628\\\textbf{(+0.015)}} \\
    \midrule
    Pretrain w jitter
      & \makecell{0.362\\raw: 0.367\\(+0.077)}
      & \makecell{0.493\\raw: 0.488\\(+0.072)}
      & \makecell{0.543\\raw: 0.559\\(+0.019)}
      & \makecell{0.610\\raw: 0.613\\(+0.010)}
      & \makecell{0.620\\raw: 0.620\\(+0.007)} \\
    \midrule
    Pretrain w flip + jitter
      & \makecell{0.380\\raw: 0.411\\(+0.095)}
      & \makecell{0.485\\raw: 0.487\\(+0.064)}
      & \makecell{0.558\\raw: 0.560\\(+0.034)}
      & \makecell{0.608\\raw: 0.616\\(+0.008)}
      & \makecell{0.621\\raw: 0.622\\(+0.008)} \\
    \bottomrule
  \end{tabular}
  \caption{\small LandCoverAI segmentation: Validation IoU at epochs 5/15/30/60/100 using a 3-epoch centered average (smoothed), with the raw per-epoch value also shown. Deltas are computed from smoothed values relative to the smoothed \emph{No pretraining}. For epoch 100 the smoothing uses the average of epochs 99 and 100. Bold deltas mark the largest improvement per column.}
  \label{tab:seg-val-iou-convergence}
\end{table}

%% file: sec/4_2_varying-downstream-data.tex
\subsection{Varying Downstream Data}
\label{subsubsec:varying-downstream-data}
To assess transferability, we finetune the best-performing DINOv2 model from Section~\ref{subsubsec:sem_seg_results} on new downstream segmentation datasets, using weights from pretraining only (not LandCoverAI segmentation finetuning). We use the LoveDA~\cite{wang2021loveda} and DeepGlobe~\cite{demir2018deepglobe} datasets for this. The model has not seen any images from these datasets during pretraining. Since DeepGlobe provides ground-truth labels only for the training split, we randomly allocate 20\% of the training data as a validation set. This split is kept fixed across all experiments.

Due to the larger size of LoveDA and DeepGlobe images, a random crop of size $512 \times 512$ is taken before any further transforms/augmentations. For validation / test images, the crop is centered for consistency. Additionally for DeepGlobe , images are first split into a grid of $4 \times 4$ tiles and saved for computational efficiency. 
\input{figures/val_iou_deepglobe_loveda_comparison}

Comparing to \textit{No pretraining} (LVD-142M weights), pretrained models show both, faster convergence and better performance (Figure~\ref{fig:val_iou_deepglobe_loveda_comparison}). Results are reported with cross-entropy loss; focal loss yielded similar trends. 

%% file: figures/val_iou_deepglobe_loveda_comparison.tex
\begin{figure}[t]
  \centering
  \includegraphics[width=\columnwidth]{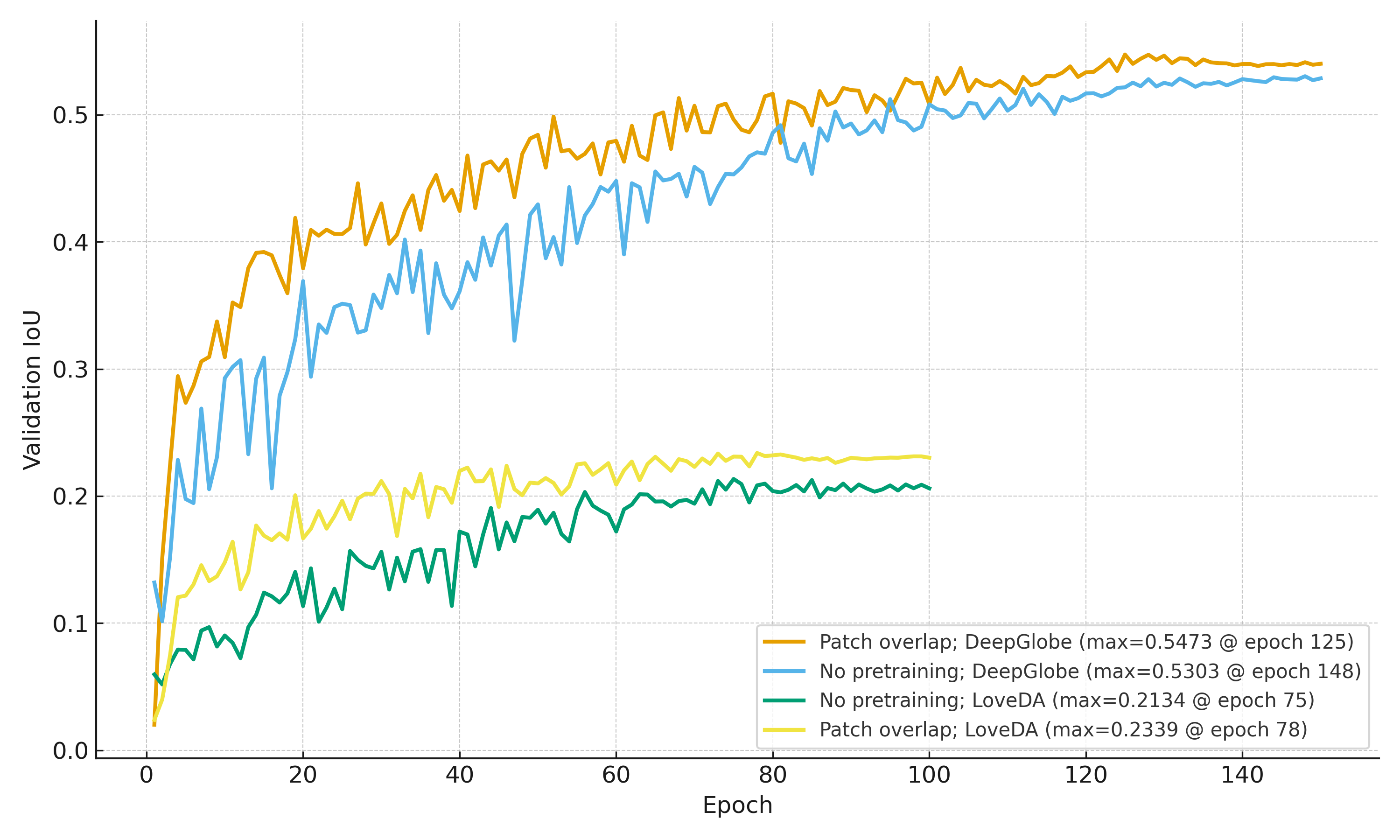}
  \caption{LoveDA and DeepGlobe segmentation: Convergence of validation IoU over training epochs. Pretrained model uses LandCoverAI for pretraining.}
  \label{fig:val_iou_deepglobe_loveda_comparison}
\end{figure}

%% file: sec/4_3_external-ssl-comparisons.tex
\subsection{External SSL Comparisons}
\label{subsubsec:external-ssl-comparisons}
Since this work is focused on efficient task-aware pretraining, we pretrain a ResNet-50 backbone and compare against pretraining methods that release ResNet-50 weights i.e. a model that can be trained on a single commodity GPU. We select methods that span a range of pretraining dataset sizes to cover different data scales. Hyperparameters are chosen based on results from Section~\ref{subsubsec:sem_seg_results}, pretraining and fine-tuning architectures are as described in Section~\ref{subsubsec:resent50-arch}.

Note that while pretraining images are resized to $224 \times 224$ (similar to DinoV2 pretraining), finetuning images for ResNet-50 were resized to $400 \times 400$.

We compare against the following:
\begin{enumerate}

    \item \textbf{GASSL:} Geography-Aware Self-Supervised Learning (GASSL)~\cite{ayush2021geography} 
    adapts MoCo-v2 to geo-tagged imagery using temporal positive pairs from spatially aligned images 
    and a geo-location prediction pretext task. It is pretrained on the Functional Map of the World 
    (fMoW) dataset~\cite{christie2018functional} with approximately 363{,}571 RGB training images and on 
    GeoImageNet, a subset of 543{,}435 geo-tagged ImageNet images~\cite{deng2009imagenet}. 
    We use weights from the MoCo-v2+Geo+TP variant.

    \item \textbf{SeCo:} Seasonal Contrast (SeCo)~\cite{manas2021seasonal} is a contrastive 
    self-supervised method for remote sensing that exploits natural seasonal and temporal 
    variations in multi-temporal Sentinel-2 data. It is pretrained on an uncurated collection 
    of Sentinel-2 data totaling about one million images. We use weights from the 
    SeCo-1M variant.

    \item \textbf{SSL4EO-S12:} Self-Supervised Learning for Earth Observation - Sentinel-1/2 (SSL4EO-S12)~\cite{wang2023ssl4eo} is a large-scale, multi-modal, 
    multi-temporal dataset for self-supervised learning in Earth observation, containing 
    approximately 3 million globally sampled Sentinel-2 images. It is used to pretrain 
    a variety of SSL approaches, including contrastive and masked-image modeling methods. 
    We use weights from the MoCo-RGB variant.

    \item \textbf{SatlasPretrain:} SatlasPretrain~\cite{bastani2023satlaspretrain} is a large-scale 
    multi-task pretraining framework built on high-resolution NAIP and Sentinel-2 imagery, with 
    hundreds of millions of labels spanning segmentation, detection, and regression tasks. 
    We use weights from the \textit{RGB Single Image} variant.

\end{enumerate}

Additionally, weights from a randomly initialized and an ImageNet pretrained backbone are used.

For all methods, including the Subimage Overlap pretrained variant, ResNet-50 backbone weights are used. These weights are loaded into the the encoder component of a U-Net segmentation model followed by finetuning on the DeepGlobe segmentation dataset.

In terms of mIoU, our method outperforms all methods except SSL4EO-S12, and is only 
marginally behind it ($0.6425$ vs.\ $0.6438$). In terms of convergence, our method 
performs best overall, reaching higher mIoU thresholds much faster than all baselines. SSL4EO-S12 — the only method with higher mIoU — is notably slow to converge. 
These results are summarized in Table~\ref{tab:ssl-thresholds-raw}.

\input{tables/ssl-best-and-thresholds}

This result is noteworthy because our method trains on substantially lesser data than the other approaches (Table~\ref{tab:dataset-scale}), yet achieves comparable or better performance while using identical architecture. To account for differences in image resolution across datasets, we also measure dataset size in terms of total pixel count, in addition to the number of images. Note that pixel count here refers only to spatial resolution, i.e. the number of spatial pixels, not pixels multiplied by the number of channels.

\input{tables/dataset-scale}

%% file: tables/ssl-best-and-thresholds.tex


\begin{table}[t]
  \centering
  \setlength{\tabcolsep}{2pt}
  \scriptsize
  \begin{tabular}{@{}>{\raggedright\arraybackslash}p{0.28\columnwidth}|
                  M{0.16\columnwidth}|*{5}{M{0.08\columnwidth}}@{}}
    \toprule
    \multicolumn{1}{c|}{Weights} & \multicolumn{1}{c|}{\begin{tabular}{@{}c@{}}Best val IoU\\(Epoch)\end{tabular}} & \multicolumn{5}{c}{Earliest epoch to reach IoU threshold} \\
    \cmidrule(lr){3-7}
    \multicolumn{2}{c|}{} & 0.60 & 0.61 & 0.62 & 0.63 & 0.64 \\
    \midrule
    Rand init                      &       0.5975 (150) &  -- &  -- &  -- &  -- &  -- \\
    \midrule
    ImageNet                       &       0.6297 (134) &  28 &  40 &  47 &  -- &  -- \\
    \midrule
    SatlasPretrain (SI)            &       0.6308 (132) &  35 &  39 &  56 & 130 &  -- \\
    \midrule
    GASSL                          &        0.6322 (97) &  19 &  28 &  39 &  97 &  -- \\
    \midrule
    SeCo                           &        0.6391 (93) &  21 &  34 &  36 &  72 &  -- \\
    \midrule
    Subimage Overlap \textit (ours)           &       0.6425 (100) &  20 &  23 &  29 &  72 &  83 \\
    \midrule
    SSL4EO\textendash S12          &       0.6438 (137) &  34 &  40 &  51 &  90 & 124 \\
    \bottomrule
  \end{tabular}
  \caption{\small Best validation IoU (raw) and earliest epoch at which the raw val IoU first reaches each threshold.}
  \label{tab:ssl-thresholds-raw}
\end{table}

%% file: tables/dataset-scale.tex
\begin{table}[t]
  \centering
  \setlength{\tabcolsep}{1.5pt}
  \scriptsize
  \begin{tabular}{@{}%
      >{\raggedright\arraybackslash}p{0.21\columnwidth}%
      >{\raggedright\arraybackslash}p{0.21\columnwidth}
      M{0.11\columnwidth}%
      M{0.11\columnwidth}%
      M{0.11\columnwidth}%
      M{0.11\columnwidth}@{}}
    \toprule
    \multicolumn{1}{c}{\makecell{Pretraining}} &
    \multicolumn{1}{c}{\makecell{Dataset}} &
    \multicolumn{1}{c}{\makecell{Images}} &
    \multicolumn{1}{c}{\makecell{Scale\\(images)}} &
    \multicolumn{1}{c}{\makecell{Pixels estimate\\(spatial)}} &
    \multicolumn{1}{c}{\makecell{Scale\\(pixels)}} \\
    \midrule
    Subimage Overlap \\(ours) & \makecell{LandCoverAI} & 10.7K & $1\times$ & 2.8B & $1\times$ \\
    \midrule
    GASSL & \makecell{FMoW RGB\\+ GeoImageNet} & 907K & $85\times$ & 54B & $19\times$ \\
    \midrule
    SeCo & \makecell{SeCo} & 1M & $94\times$ & 70B & $25\times$ \\
    \midrule
    SSL4EO\textendash S12 & \makecell{SSL4EO\textendash S12} & 3M & $281\times$ & 209B & $75\times$ \\
    \midrule
    SatlasPretrain & \makecell{SatlasPretrain} & 856K & $80\times$ & 3.3T & $1180\times$ \\
    \bottomrule
  \end{tabular}
  \caption{\small Dataset scale comparison for pretraining. Relative sizes (scale) computed w.r.t.\ \emph{LandCoverAI}.}
  \label{tab:dataset-scale}
\end{table}

%% file: sec/5_conclusion.tex
\section{Conclusion}
\label{sec:conclusion}
Self-supervised learning is valuable in remote sensing, where imagery is abundant but labels are costly. Since most existing approaches train foundation models on large scale data, we study how smaller, task-aligned pretraining can provide efficiency with improved downstream performance.

We introduce Subimage Overlap Prediction as a new spatial auxiliary task: given a full image and a random subimages selected from it, the model predicts the subimage’s location as a semantic mask. We show that task-aware pretraining using Subimage Overlap Prediction improves downstream land-cover segmentation over standard initialization (ImageNet, LVD-142M) and transfers to datasets that differ from the pretraining distribution, with increasing gains as labeled data is reduced. Despite using far less pretraining imagery, our method matches or surpasses recent SSL baselines. Useful future directions include applying this Subimage Overlap to other dense prediction tasks that are common in remote sensing 
(e.g., object detection, change detection), investigating how performance scales with 
larger pretraining datasets, and more broadly exploring task-aligned pretraining methods for specific downstream tasks.